\documentclass{article}

\usepackage[final]{neurips_2019}

\usepackage[utf8]{inputenc} 
\usepackage[T1]{fontenc}    
\usepackage{hyperref}       
\usepackage{url}            
\usepackage{booktabs}       
\usepackage{amsfonts}       
\usepackage{nicefrac}       
\usepackage{microtype}      
\usepackage{graphicx}
\usepackage{floatrow}   
\usepackage{subcaption}

\title{Risks of Using Non-verified Open Data: A case study on using Machine Learning techniques for predicting Pregnancy Outcomes in India}

\author{Anusua Trivedi$^{*}$, \textbf{Sumit Mukherjee$^{*}$,} \textbf{Edmund Tse$^{*}$,} \textbf{Anne Ewing,} \textbf{Juan Lavista Ferres}\\ 
Microsoft, Redmond, USA. \\
$^{*}$ These authors contributed equally.
}
\begin{document}
\maketitle
\vspace{-.5cm}
\begin{abstract}
\vspace{-0.3cm}
  Artificial intelligence (AI) has evolved considerably in the last few years. While applications of AI is now becoming more common in fields like retail and marketing, application of AI in solving problems related to developing countries is still an emerging topic. Specially, AI applications in resource-poor settings remains relatively nascent. There is a huge scope of AI being used in such settings. For example, researchers have started exploring AI applications to reduce poverty and deliver a broad range of critical public services. However, despite many promising use cases, there are many dataset related challenges that one has to overcome in such projects. These challenges often take the form of missing data, incorrectly collected data and improperly labeled variables, among other factors. As a result, we can often end up using data that is not representative of the problem we are trying to solve. In this case study, we explore the challenges of using such an open dataset from India, to predict an important health outcome. We highlight how the use of AI without proper understanding of reporting metrics can lead to erroneous conclusions.
\vspace{-0.3cm}
\end{abstract}
\vspace{-0.3cm}
\section*{Introduction}
\vspace{-0.3cm}
An estimated 2.6 million still births and 2.7 million neo-natal deaths occur globally each year [8]. A disproportionate percentage (approximately 99\%) of these deaths occur in low-income and middle-income developing countries [9]. India accounts for approximately 19 births/1,000 population (2017 est.) and 39.1 deaths/1,000 live births (2017 est.)[16]. While India has made laudable progress in reducing overall child mortality over the past 25 years, similar reductions in neonatal mortality have lagged. As a result, India fell short of the Millennium Development Goal 4 of a two-thirds reduction in under-five mortality from 1990 to 2015 [10]. Previous research on evidence-based approaches to reducing infant mortality have included efforts to increase access to preventive and curative antenatal, perinatal and postnatal care [11]. Pregnancy outcomes is a strong indicator of such cares [12]. Unfortunately there is a big divide between the developed and developing world on this pregnancy outcome. This is not because of a lack of clarity in defining a pregnancy outcome, but because of an inherent weakness in the health information and recording systems [17]. According to Indian Council of Medical Research (ICMR) [13], data on health and demographics in India is plagued by incomplete information, overestimation, and over-reporting that lead to hindrance in policy planning. The National Data Quality Forum (NDQF) [14], formulated by ICMR’s National Institute for Medical Statistics (ICMR - NIMS), in partnership with Population Council has identified gaps in data compilation. The NDQF attempted to identify issues in data quality, and found several factors effecting the data quality:
\begin{itemize}
\item lack of comparability 
\item poor usability of national level data sources
\item discordance between system and survey level estimates
\item increased questionnaire length
\item capturing wrong subsections for questions in questionnaire
\item questions on socially restricted conversation topics
\item age-reporting errors
\item non-response and intentional skipping of questions
\item under reporting due to subjective question interpretation
\item incompleteness of data
\end{itemize}
All these factors translate to poor data quality. For our case study we explore open reporting data sources in India and try to show how use of AI without understanding these reporting can lead to wrong estimates.
unexplored in our experiment.
\vspace{-0.3cm}
\section*{Data}
\vspace{-0.3cm}
\paragraph{Pregnancy Outcome Data} For calculating pregnancy outcomes in India, we started exploring two major data sources: Open Government Data (OGD) platform [2] and Kaggle [3]. OGD is a platform for supporting Open Data initiative of Government of India. The portal is used by Govt. of India to publish datasets, documents, services, tools and applications collected for public use. It intends to increase transparency in the functioning of Government and also open avenues for many more innovative uses of Government Data to give different perspective. The base Open Government Data Platform India is a joint initiative of Government of India and US Government. Kaggle is a crowd-sourced platform to attract, nurture, train and challenge data scientists from all around the world to solve data science, machine learning and predictive analytics problems. Kaggle enables data scientists and other developers to engage in running machine learning contests, write and share code, and to host datasets. Kaggle users have shared over one thousand datasets and is the number one go-to place for AI practitioners to look for data [4]. For our case study we looked at the Annual Health Survey (AHS) on Woman Health for 9 states in India [5]. Specifically, we focused on the \textit{Woman} dataset and \textit{WPS} dataset in [5]. The \textit{Woman} dataset is also represented in Kaggle [6] with acknowledgement to the source being OGD Platform. The data curator of the Kaggle pregnancy outcome dataset states - "These nine states, which account for about 48 percent of the total population, 59 percent of Births, 70 percent of Infant Deaths, 75 percent of Under 5 Deaths and 62 percent of Maternal Deaths in the country, are the high focus States in view of their relatively higher fertility and mortality". These numbers did not make any sense to us, so we started to explore this dataset further. 
\paragraph{Data Gap Analysis} We came across the AHS questionnaire form used by the Census Office in India [7]. This questionnaire was used to collect the pregnancy outcome dataset in [5]. Further exploration of this questionnaire form showed us some interesting facts which helped us identify some data gaps. 
\begin{figure}[!h]
    \centering
    \fbox{\includegraphics[scale=0.6]{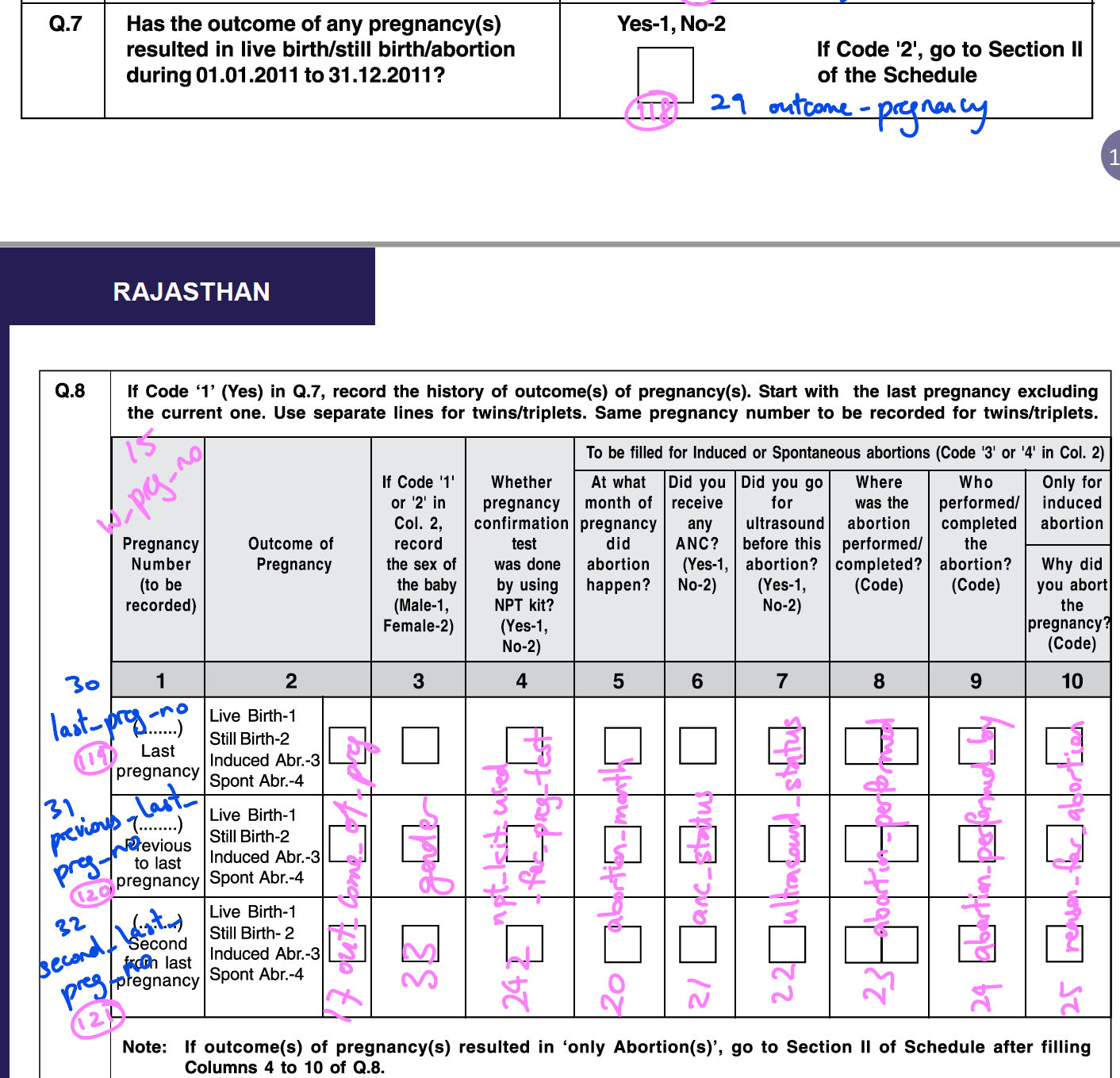}}
    \caption{Snapshot of the AHS questionnaire for the state of Rajasthan in India}
    \label{fig1}
\vspace{-0.3cm}
\end{figure}
Figure~\ref{fig1} shows a snapshot of the AHS questionnaire for the state of Rajasthan in India. In Figure~\ref{fig1}, if the answer to the Question 7 is \textit{Yes}, that answer leads to Question 8, and if the answer to the question 7 is \textit{No}, that answer leads to Question 9. But this is not correctly represented in the \textit{Woman} dataset being used by Kaggle in [6]. In [6], the Question 8 data is completely missing. So the \textit{pregnancy outcome} column in Kaggle dataset [6] comprises of just binary pregnancy outcome labels coming from Question 7 (\textit{Yes}, \textit{No}), instead of 4 pregnancy outcome labels coming from Question 8 (\textit{Live Birth}, \textit{Still Birth}, \textit{Induced Abortion}, \textit{Spontaneous Abortion}). The \textit{pregnancy outcome} column in the \textit{WPS} dataset on the other hand comprises of 4 pregnancy outcome labels coming from Question 8 (\textit{Live Birth}, \textit{Still Birth}, \textit{Induced Abortion}, \textit{Spontaneous Abortion}). This data gap leads us to believe that the \textit{Woman} dataset should not be used to train AI models for pregnancy outcome prediction. That makes the Kaggle dataset in [6] unfit for training such a model. We should rather be using \textit{WPS} dataset to train AI models for pregnancy outcome prediction. To evaluate our hypothesis, we run some experiments on both  \textit{Woman} and \textit{WPS} dataset.
unexplored in our experiment.
\vspace{-0.3cm}
\section*{Experiment}
\vspace{-0.3cm}
\begin{figure}[h!]
    \centering
    \includegraphics[scale=0.65]{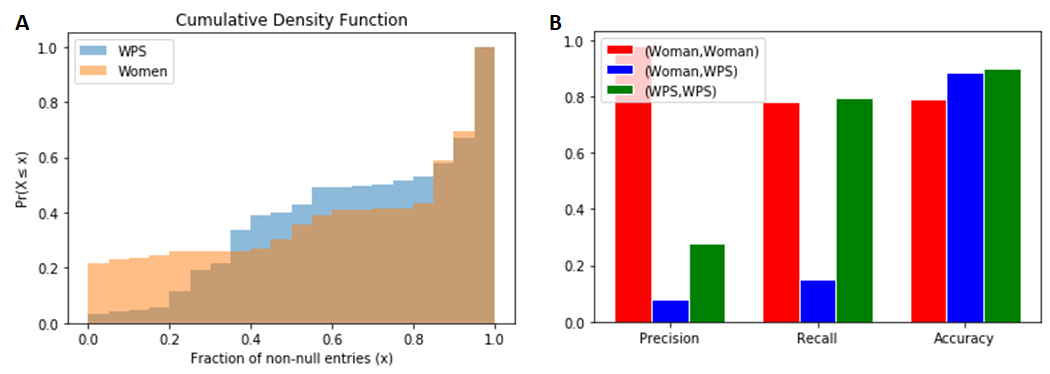}
    \caption{a) Cumulative density function of fraction of null entries of features in the \textit{WPS} and \textit{Woman} datasets. b) Comparison of models trained on \textit{Woman} and \textit{WPS} datasets - we see that women trains with the Women dataset leads to poor performance on the \textit{WPS} data in comparison to models trained on \textit{WPS} data. }
    \label{fig2}
\vspace{-0.3cm}
\end{figure}
To demonstrate the problems with using incorrect labels, we sought to demonstrate how models trained with the incorrect and correct labels are different from each other. To this end, we first sought to identify usable features from each dataset. The \textit{Women} dataset had 202 features, while \textit{WPS} had 355 features. The features in the \textit{Woman} dataset were a sub-set of those in the \textit{WPS} dataset. Furthermore, the \textit{WPS} dataset is supposed to only contain individuals who responded \textit{Yes} to the \textit{pregnancy-outcome} question. To keep the models comparable, we only consider the features that are available in both datasets. We further down the features to discard all features that were not related to social, economic, health or demographic indicators e.g. all \textit{id} related fields were discarded. Since most confounding features which captured birth outcome information (e.g. \textit{is-receive-birth-certification}, \textit{baby-checkup-after birth-in-days} etc.). Since the outcome variable for the \textit{WPS} dataset was not a binary variable we converted it into a binary survival prediction problem by treating \textit{Live birth surviving} as one category and everything else as the \textit{Non-surviving} category. For both datasets, we discarded entries for which the outcome variable was null. As a further step in the data cleaning process, we discarded all features that had over 10 percent entries as null. We noticed an extremely high number of null entries in both datasets as can be seen in Figure~\ref{fig2}a, which highlights a significant problem in the data collection process. We further proceeded to one-hot-encode all categorical variables present in the datasets. Having performed the data cleaning/processing, we split each processed dataset into a 2:1 training/test split. At this stage both datasets have the same features and we train separate logistic regression models on both training sets using the scikit-learn python package. We then evaluate the learned models on: i) the test set corresponding to their training set, ii) the test set corresponding to the other dataset (Figure~\ref{fig2}b). We see a large drop in precision/recall/accuracy for the model trained on the \textit{Woman} dataset and tested on the \textit{WPS} dataset in comparison to the model trained on the \textit{WPS} dataset. Moreover, the model trained on the \textit{Woman} performs extremely well on the test data for \textit{Woman}. Despite the extremely good performance on many of the metrics, this model is actually prediction whether or not the woman was pregnant based on the available feature information and not what the outcome of the pregnancy was. This model was in fact a rather poorly performing model for the task of predicting pregnancy outcomes compared to a model actually trained to predict pregnancy outcomes. This highlights a one of the major issues with using such datasets, which is the lack of clarity in naming convention which can lead to misinterpretation of features and outcomes. While a long term solution to this problem is adoption of clear and consistent naming conventions for variables, a shorter term solution is for users of such datasets to rigorously explore the data dictionaries and survey forms. It must be noted that there may have been many other types of reporting errors which were unexplored in our experiment.
\vspace{-0.3cm}
\section*{Discussion}
\vspace{-0.3cm}
Due to the complicated textual nature of the AHS questionnaire, we end up capturing wrong subsections for questions in questionnaire for our case study. By taking steps to simplify current AHS forms and making sure to correctly transition to online data, we will have a much more comprehensive data set to draw meaningful conclusions from. By doing so, we can ensure that we are applying AI in a more efficient manner to predict pregnancy outcomes. Despite the highlighted issues in the data collection and processing of such datasets from developing countries, there is still a lot of value in analyzing such datasets. While this is beyond the scope of this paper, the \textit{WPS} dataset, once appropriately processed, could be used to build an accurate pregnancy outcome prediction model. In our simple experiment, we already saw decent recall and accuracy for the model trained and tested on \textit{WPS}. Such a model, along with analysis of the important predictors would help identify women whose pregnancies are at risk as well as the socio-economic or healthcare factors that are leading to adverse pregnancy outcome in such developing nations.
unexplored in our experiment.
\vspace{-0.3cm}
\section*{References}
\vspace{-0.4cm}
\medskip
\small
[1] WHO, UNICEF, UNFPA. Maternal mortality in 2000: Estimates developed by WHO, UNICEF, UNFPA. Geneva: WHO, 2003

[2] Open Data Government Platform India: https://data.gov.in

[3] Kaggle: https://www.kaggle.com

[4] Kaggle Getting Started: https://www.kaggle.com/getting-started/44916

[5] Annual Health Survey - Woman Schedule: https://data.gov.in/catalog/annual-health-survey-woman-schedule

[6] Kaggle Predict Outcome of Pregnancy: https://www.kaggle.com/rajanand/ahs-woman-1

[7] Ministry of Home Affairs, Government of India:\\ http://www.censusindia.gov.in/vital\_statistics/AHSBulletins/Factsheets.html

[8] UNICEF Data: monitoring the situation of children and women:\\
http://data.unicef.org/child-mortality/neonatal.html

[9] World Health Organization (WHO). The partnership for maternal, newborn and child health.\\http://www.who.int/pmnch/media/press\_materials/fs/fs\_newborndealth\_illness/en/

[10] NIMS, ICMR and UNICEF. Infant and child mortality in india: levels, trends and determinants. New Delhi, India: NIMS, ICMR and UNICEF 2012. http://unicef.in/CkEditor/ck\_Uploaded\_Images/img\_1365.pdf

[11] Koblinsky M , Matthews Z , Hussein J , et al . Lancet maternal survival series steering group. Going to scale with professional skilled care. Lancet 2006;368:1377–86.

[12] Strehlow MC , Newberry JA , Bills CB , et al . Characteristics and outcomes of women using emergency medical services for third-trimester pregnancy-related problems in India: a prospective observational study. BMJ Open 2016;6:7.doi:10.1136/bmjopen-2016-011459 

[13] Indian Council of Medical Research: https://www.icmr.nic.in/

[14] National Data Quality Forum: https://www.insightsonindia.com/2019/07/26/national-data-quality-forum-ndqf/

[15] National Institute for Medical Statistics: http://nims-icmr.nic.in/NIMS/index.jsp

[16] India Demographics Profile 2018: https://www.indexmundi.com/india/demographics\_profile.html

[17] The Challenges Confronting Public Hospitals in India, Their Origins, and Possible Solutions:\\https://www.hindawi.com/journals/aph/2014/898502/
\end{document}